\documentclass[10pt,twocolumn,letterpaper]{article}

\usepackage{cvpr}              

\usepackage{graphicx}
\usepackage{amsmath}
\usepackage{amssymb}
\usepackage{booktabs}
\usepackage[pagebackref,breaklinks,colorlinks]{hyperref}

\usepackage[capitalize]{cleveref}
\crefname{section}{Sec.}{Secs.}
\Crefname{section}{Section}{Sections}
\Crefname{table}{Table}{Tables}
\crefname{table}{Tab.}{Tabs.}

\def\cvprPaperID{*****} 
\def\confName{CVPR}
\def\confYear{2022}

\begin{document}

\title{Multi-View 3D Reconstruction using Knowledge Distillation}


\author{
\begin{tabular}{ccc}
\textbf{Aditya Dutt*} & \textbf{Ishikaa Lunawat*} & \textbf{Manpreet Kaur*} \\
{\tt\small asdutt2@stanford.edu} &
{\tt\small ishikaa@stanford.edu} &
{\tt\small mkaur5@stanford.edu}
\end{tabular}
\\[0.8em]
Stanford University
\\[0.1em]
{\footnotesize \textit{* Equal contribution.}}
}

\maketitle

\begin{abstract}
Large Foundation Models like Dust3r can produce high quality outputs such as pointmaps, camera intrinsics, and depth estimation, given stereo-image pairs as input. However, the application of these outputs on tasks like Visual Localization requires a large amount of inference time and compute resources. To address these limitations, in this paper, we propose the use of a knowledge distillation pipeline, where we aim to build a student-teacher model with Dust3r as the teacher and explore multiple architectures of student models that are trained using the 3D reconstructed points output by Dust3r. Our goal is to build student models that can learn scene-specific representations and output 3D points with replicable performance such as Dust3r. The data set we used to train our models is 12Scenes. We test two main architectures of models: a CNN-based architecture and a Vision Transformer-based architecture. For each architecture, we also compare the use of pre-trained models against models built from scratch. We qualitatively compare the reconstructed 3D points output by the student model against Dust3r's and discuss the various features learned by the student model. We also perform ablation studies on the models through hyperparameter tuning. Overall, we observe that the Vision Transformer presents the best performance visually and quantitatively.   
\end{abstract}

\section{Introduction}
\label{sec:intro}

Obtaining multi-view 3D reconstruction from 2D images is a challenging task. Recently, this has been made easier through the use of large foundation models that can be trained on large sets of data. An example of a foundation model is DUSt3R \cite{wang2023dust3r}. DUSt3R aims to solve the multiview reconstruction problem for stereoimage pairs through direct scene coordinate regression without making any prior assumptions on camera intrinsics or extrinsic parameters. The strength of DUSt3R is observed in its applications towards downstream tasks like Visual Localization and 3D Reconstruction. For Visual Localization, the model first performs Pixel Correspondence matching and is then evaluated on two different datasets, performing comparably well on unseen images, despite having not been trained on visual localization tasks at all. However, a lot of processing time is required for these tasks, as the model only works with stereo-image pairs. Moreover, the 3D points are not output in the world reference frame. In order to address these issues, we plan to build a smaller neural network model that learns from the pre-trained foundation model through a knowledge distillation framework and outputs 3D points relative to a fixed world coordinate system. The smaller network will be trained to learn scene-specific knowledge and will be faster and more lightweight than DUSt3R.


\section{Related Work}
\label{sec:relatedwork}
\paragraph{3D Dense Reconstruction} 

Papers such as Accelerated Coordinate Encoding \cite{brachmann2023accelerated} propose a light-weight and ultra-fast neural network that predicts 3D coordinates for every pixel in an image. This paper establishes an important benchmark for neural network models that can be trained for a specific scene in under 5 mins and can predict a point cloud associated with an image in real-time. While the time to train a scene coordinate network is quite less, training such networks in real-time on mobile devices is often impractical due to lack of compute and high memory consumption. Hence, we look at an advent shift in paradigm brought on by foundation models.

The paper introduced by \cite{wang2023dust3r} introduces a model that predicts 3D location corresponding to every pixel in the image without any scene-specific training, giving this method a remarkable ability to generalize to any scene. This paper then extends their sparse point cloud prediction model to solve downstream 3D computer vision tasks such as relative pose estimation, monocular depth estimation, etc and outperforms previous state of the art works. 

\paragraph{Knowledge-Distillation}
Knowledge distillation has been widely applied in the field of convolutional neural networks (CNNs) and vision models to enhance their efficiency and performance. In applications such as image classification, object detection, and semantic segmentation, distillation techniques enable the deployment of lightweight student models that maintain high accuracy while reducing computational load . Studies such as \cite{hinton2015distilling} have shown that knowledge distillation can improve the performance of compact models in real-time vision tasks, making it particularly valuable for resource-constrained environments like mobile and embedded devices . Advanced methods like attention transfer by \cite{zagoruyko2017paying} and intermediate layer guidance by \cite{romero2015fitnets} further refine the distillation process, ensuring that student models capture critical features and patterns from their teachers, thereby optimizing their performance in diverse vision applications

\section{Problem Statement}
\label{sec:probstatement}
    To address some of the shortcomings of the DUSt3R model, we plan to implement a smaller neural network that learns scene-specific information and provides generalized 3D reconstructed points expressed in a fixed world coordinate system, which will improve tasks such as Visual Localization. Details about the model architecture and our overall method setup is described in Section \ref{sec:techapproach}. The dataset we will use is the \textit{12Scenes} dataset that stores rich scene-specific RGB-D data of 4 large scenes, containing 12 rooms. During training, we will evaluate our model using \textit{Mean-Square Loss} (MSE) to increase the accuracy of the predicted 3D point locations. With the help of the knowledge distillation framework and data containing scene-specific information, we hope that our network predicts accurate world 3D points.

\section{Approach}
\label{sec:techapproach}


\subsection{Knowledge-Distillation}

For our problem, we propose a knowledge distillation framework consisting of a teacher and student model. Our student model is designed as a convolutional neural network (CNN) and a standard vision transformer (ViT) that learns from our large teacher model, Dust3R \cite{wang2023dust3r}. The code for our approach is provided here, \href{https://github.com/ishikaalunawat/231aproj}{Github}. Our approach is as follows: 

\begin{enumerate}
    \item \textbf{Dataset Preparation}: 
    We use few scenes from the \textit{12Scenes} dataset \cite{valentin2016learning} and create pairwise images with intersecting views to serve as input for the Dust3R model. This involves loading images and pairing them based on overlapping views of the scene.
    \item \textbf{Teacher Model Inference}:
    The Dust3R model generates 3D coordinates for all pairs of images. These coordinates are obtained through an inference process and are used as the ground truth labels for training the Student model. Note that Dust3R provide 3D points in the frame of reference of the first image.
    \item \textbf{Global Alignment Step}:
    To ensure consistency, we perform a global alignment process to align and transform the predicted 3D points to the same frame of reference. This involves fixing an origin point and aligning all predicted points accordingly, and is performed as a post-processing step.
    \item \textbf{Training Student Model}:
    Using the aligned 3D points from the Dust3R model as labels, we train the Student model. The training objective is to minimize the MSE loss between the predicted 3D points from the Student model and the labels provided by the Dust3R Teacher model. As our preliminary test, we design the network as a six layer CNN, where each layer is followed by ReLU.
    
\end{enumerate}

\subsection{Student Models}
\begin{figure}[!h]
  \centering
  \begin{subfigure}{\linewidth}
    \includegraphics[width=\linewidth]{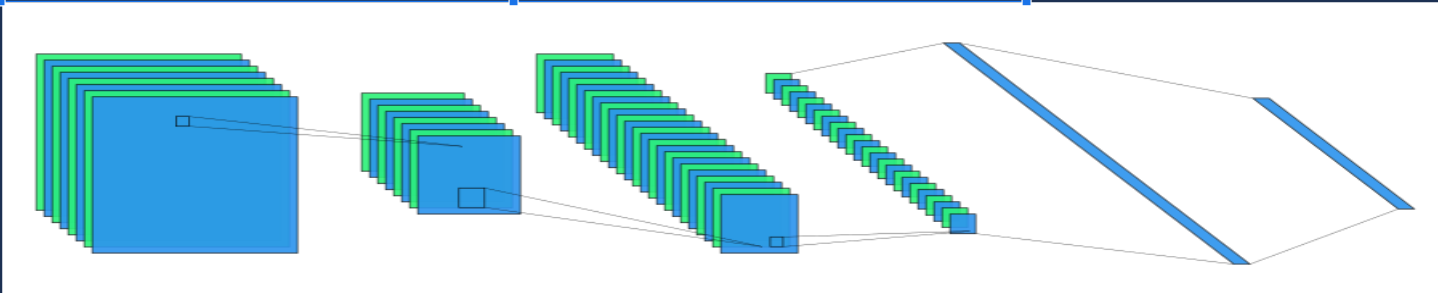}
    \caption{Vanilla CNN Architecture}
  \end{subfigure}
  
  \vspace{0.5cm}
  
  \begin{subfigure}{\linewidth}
    \includegraphics[width=\linewidth]{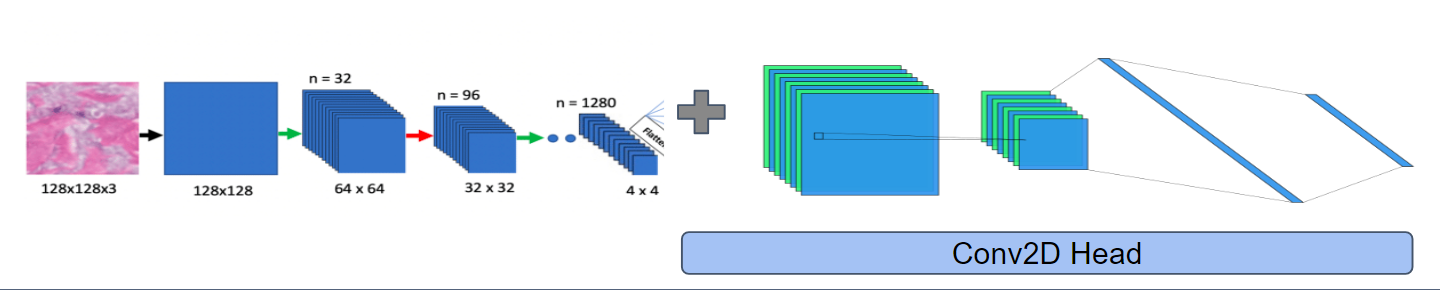}
    \caption{MobileNetV3 Model with a Conv2D Head}
  \end{subfigure}
  
  \vspace{0.5cm}
  
  \begin{subfigure}{\linewidth}
    \includegraphics[width=\linewidth]{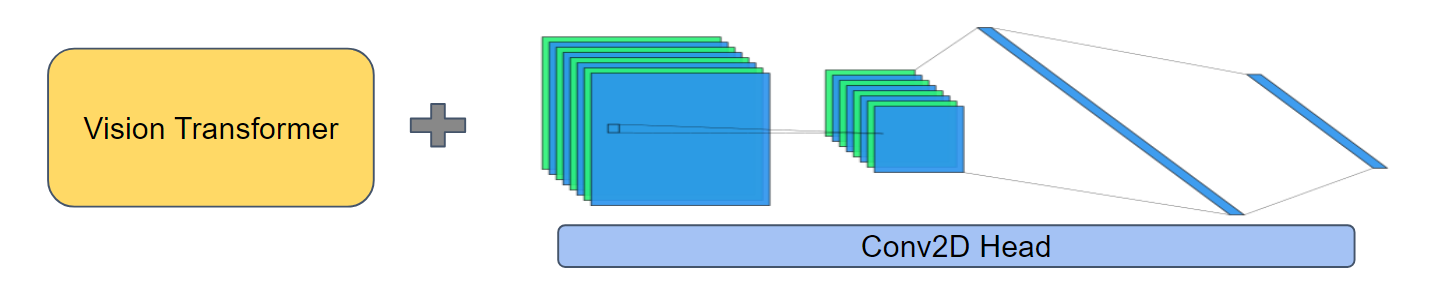}
    \caption{Vision Transformer Architecture with a Conv2D Head}
    \label{fig:short-c}
  \end{subfigure}
  \caption{Reconstructed Kitchen scene with camera poses using DUST3R model and global optimization method}
  \label{fig:student_models}
\end{figure}

We primary explored two types of model architectures: a CNN-based model and a Vision Transformer based model. For the CNN architecture, we compared the use of a vanilla model structure against the use of an existing pre-trained model that encodes features and attaching a convolutional head to its tail. A detailed description of each of the student models is provided below. Figure \ref{fig:student_models} also depicts the different types of models.

\begin{enumerate}
    \item \textbf{Vanilla CNN}: This model implements a simple convolutional neural network that takes in 3-channel RGB images and scales it up to 512 channels. At the end, a set of fully-connected layers are used to output the 3D points for each pixel in the RGB Image. The output layers (that are fully-connected) is repeated across all the different models. This network is 45MB in size, and hence can be considered light-weight and perfect for edge deployment.

    \item \textbf{Pre-trained MobileNetV3 Model with a Conv Head}: This model uses a pre-trained version of the model developed in \cite{howard2019searching}. A Conv Head at the end of the model replaces the classification head used in the original Mobilenet model in \cite{howard2019searching} in order to output 3D points for each pixel in our image. This Convolutional Head is replicated as the output head for all of our student models as is described in each section. This network is 3.7MB in size. Even though it's much smaller than the Vanilla CNN network, the performance does not degrade compared to Vanilla CNN as noted in the Results section. 

    \item \textbf{Vision Transformer}: 
    The Vision Transformer consists of Encoder + Decoder layers which are based on the paper \cite{vit} and described in detail below.

    \textit{Patch Extractor}
The Patch Extractor module divides the input image \( x \) into non-overlapping patches using unfolding operations.

\textit{Input Embedding}
The Input Embedding module projects each patch into a latent space and adds positional embeddings. The final embedding combines the class token and the linear projections of the patches with positional embeddings.

\textit{Encoder Block}
Each Encoder Block employs a standard Transformer encoder structure consisting of Layer Normalization, Multi-Head Attention, and a Multi-Layer Perceptron (MLP) with GELU activation and dropout. The residual connections around the attention and MLP layers are commonly used for better the gradient flow and model performance.

\textit{Decoder Block}
The Decoder Block mirrors the Encoder Block's architecture, bringing the transformation of latent representations back into a form suitable for the final convolutional head. It uses similar components: Layer Normalization, Multi-Head Attention, and an MLP.

\textit{Convolutional Head}
The Convolutional Head processes the reshaped output from the decoder, applying a series of convolutional layers to generate the final output. The head includes Leaky ReLU activations between convolutional layers. The final output is a point for every pixel or a pointcloud.

\end{enumerate}


\section{Results}
\label{sec:results}

\begin{figure}[!h]
    \centering
    \begin{subfigure}{0.49\linewidth}
        \includegraphics[width=\linewidth]{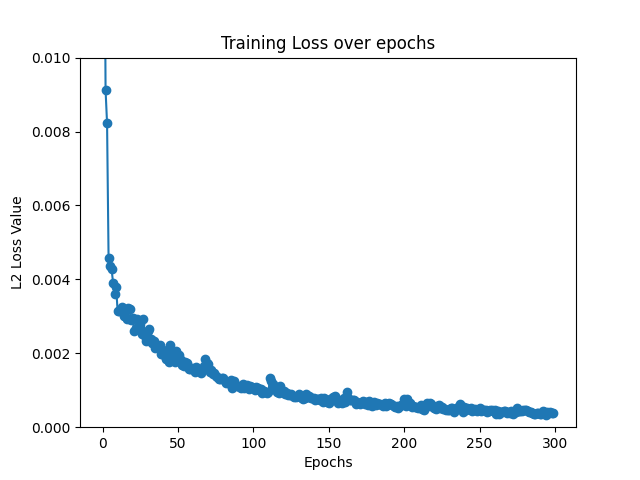}
        \caption{Training loss vs. epochs for Kitchen scene with Vanilla CNN}
        \label{fig:kitchen-loss}
    \end{subfigure}
    \hfill
    \begin{subfigure}{0.49\linewidth}
        \includegraphics[width=\linewidth]{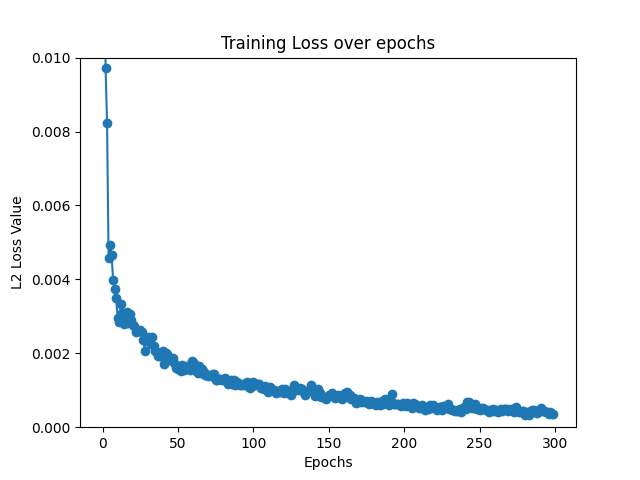}
        \caption{Training loss vs. epochs for Office scene with MobileNet CNN}
        \label{fig:office-loss}
    \end{subfigure}
    \begin{subfigure}{0.49\linewidth}
        \includegraphics[width=\linewidth]{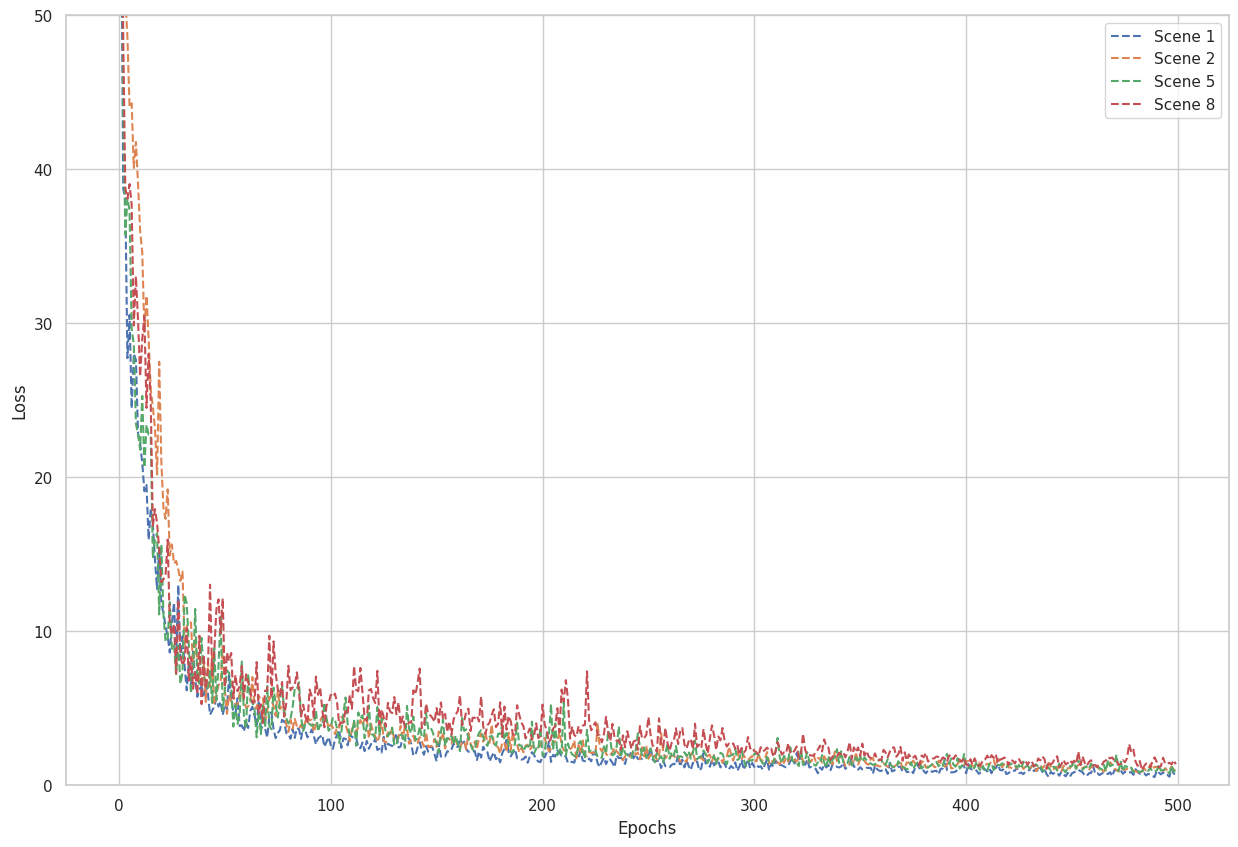}
        \caption{Training loss vs. epochs for multiple scenes with Vision Transformer architecture}
        \label{fig:vit-loss}
    \end{subfigure}
    \caption{Training loss vs. epochs for different scenes}
    \label{fig:training-loss}
\end{figure}

\begin{figure}[!h]
  \centering
  \begin{subfigure}{0.46\linewidth}
    \includegraphics[width=\linewidth]{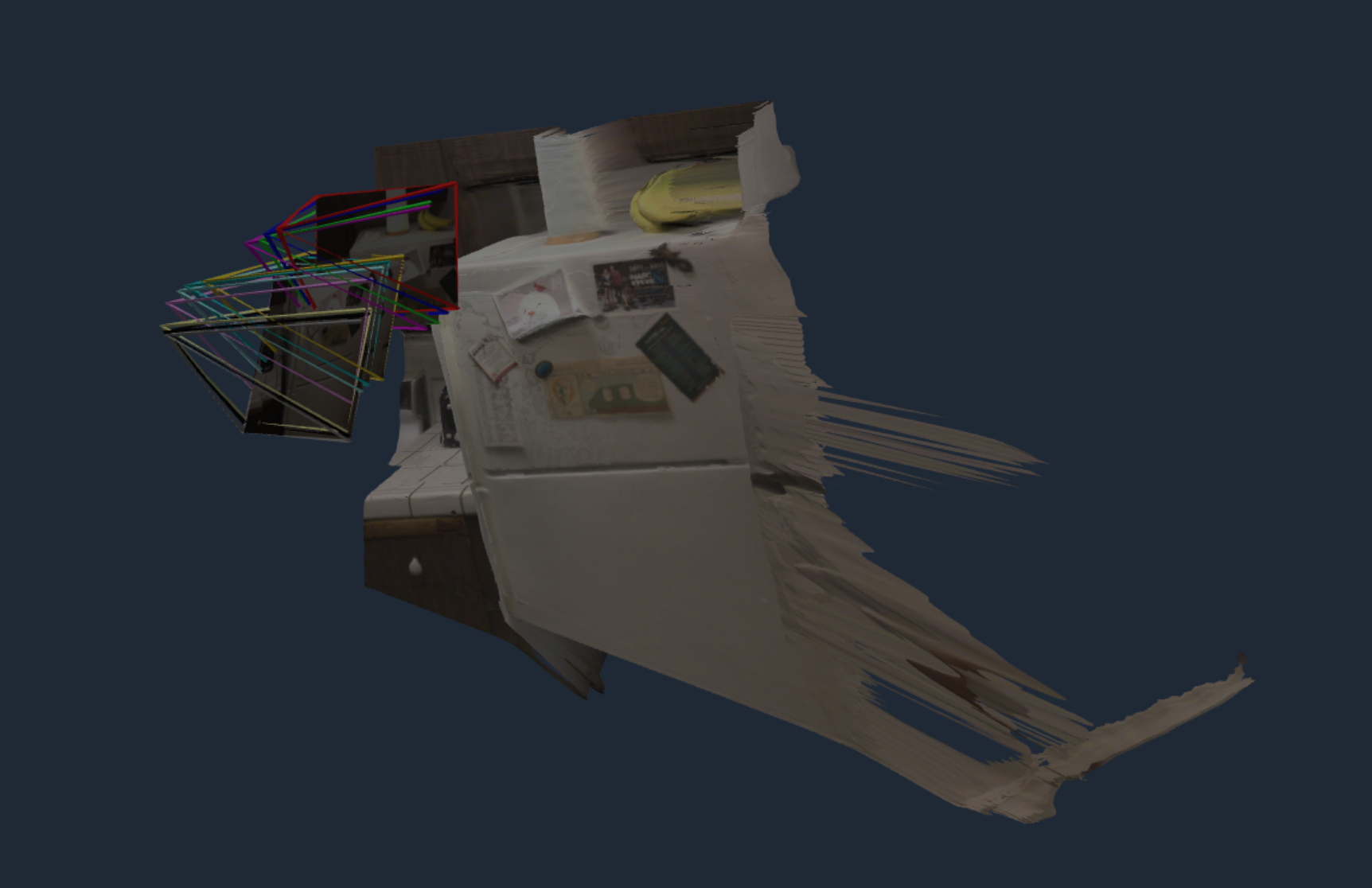}
    \caption{Front-view of kitchen scene}
    \label{fig:short-a}
  \end{subfigure}
  \hfill
  \begin{subfigure}{0.46\linewidth}
    \includegraphics[width=\linewidth]{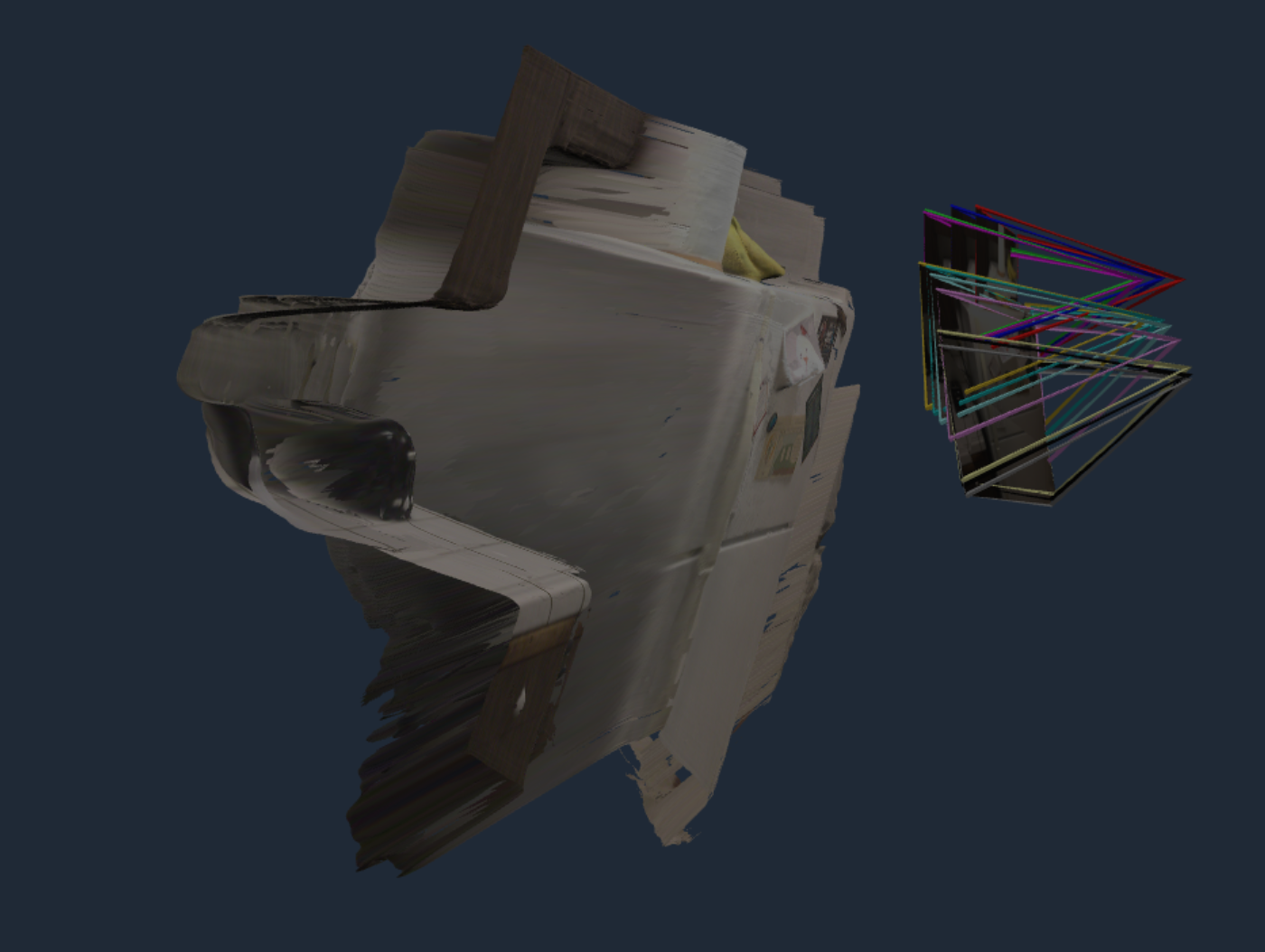}
    \caption{Side-view of kitchen scene}
    \label{fig:short-b}
  \end{subfigure}

  \vspace{0.5cm}

  \begin{subfigure}{0.46\linewidth}
    \includegraphics[width=\linewidth]{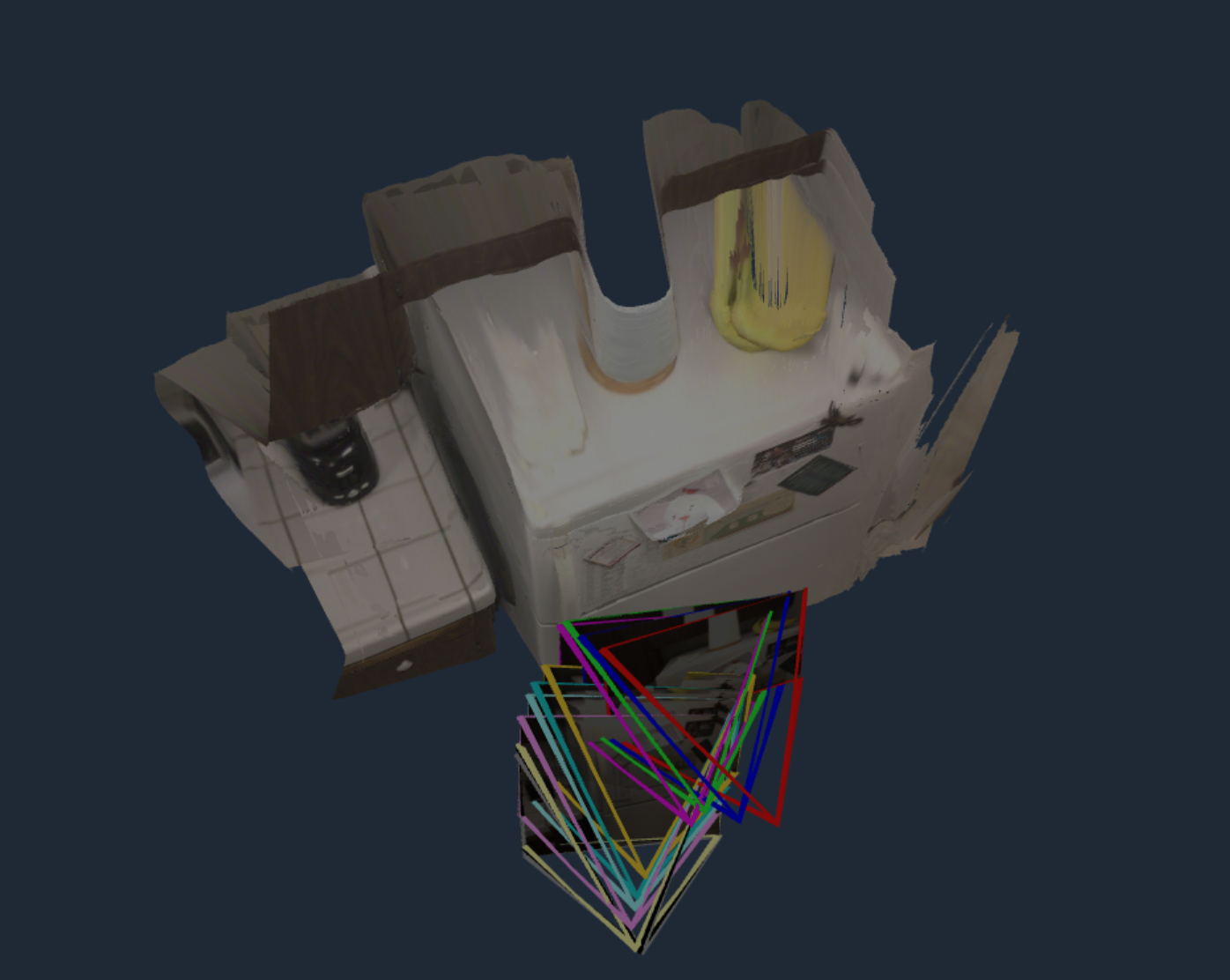}
    \caption{Top-view of the kitchen scene}
    \label{fig:short-c}
  \end{subfigure}
  \hfill
  \begin{subfigure}{0.46\linewidth}
    \includegraphics[width=\linewidth]{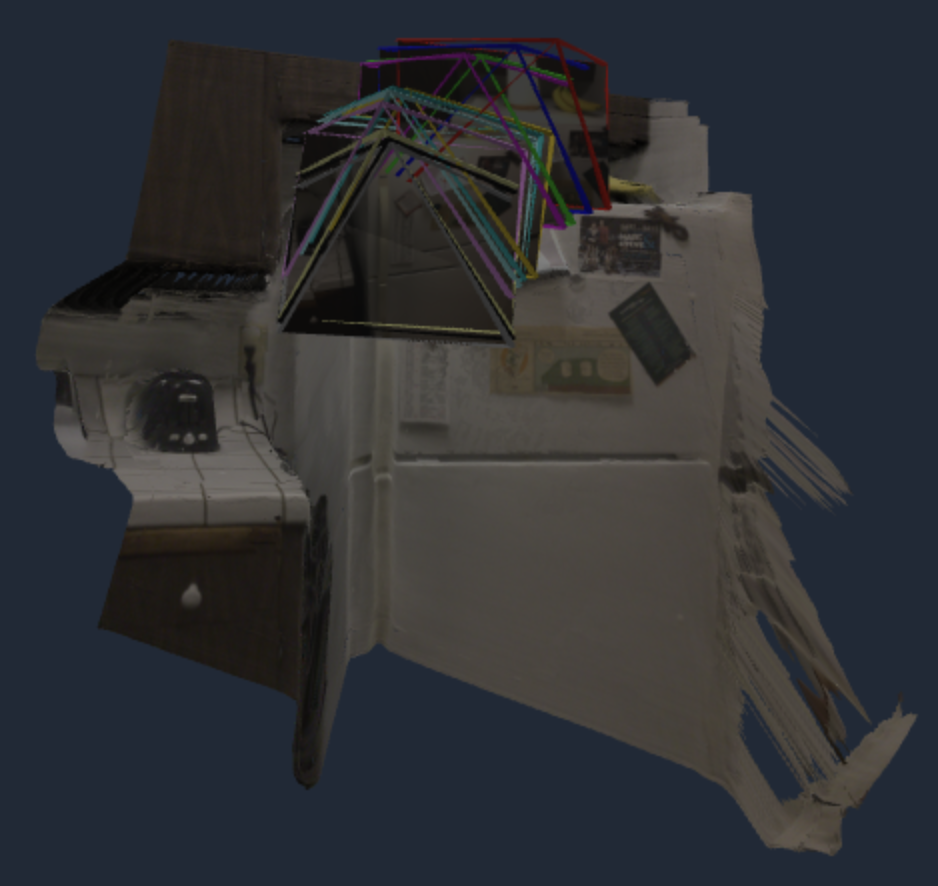}
    \caption{Side-view of the kitchen scene}
    \label{fig:short-d}
  \end{subfigure}

  \caption{Reconstructed Kitchen scene with camera poses using DUST3R model and global optimization method}
  \label{fig:3d-kitchen}
\end{figure}

\begin{enumerate}

\item \textbf{3D Recon results using DUST3R} 
The DUST3R model can be used to predict 3D coordinates for a pair of images. However, in order to reconstruct a scene with more than 2 images, we first do pairwise inference of each image and then perform a global alignment over all pairwise predictions to obtain world-coordinate pointmaps for all the images. We show the results for a small part of the Kitchen scene in Apartment 1 from the 12scenes dataset in Figure \ref{fig:3d-kitchen}.

\item \textbf{Training Student Model Training Loss}: We set up a knowledge distillation training pipeline to use the Teacher Model (DUST3R) for predicting 3D coordinates and use the predicted 3D coordinates to learn the Student Model. 


We see a general downward trend of training loss which converges to around 0.00037 for kitchen and 0.0004 for office space, as can be seen in Figures \ref{fig:kitchen-loss} and \ref{fig:office-loss} for the Vanilla CNN and MobileNet architectures. We also show the downward trend for Vision Transformer training architecture in \ref{fig:vit-loss}. We would like to note that we trained the CNN networks in the original scale of DUST3R's output. However, we change the scale by multiplying it by 100 for vision transformer model training. This was just to test numerical stability of the networks. We don't think this scaling has an impact on learing as both training losses trend downward by very similar orders.

\item \textbf{Mean L2 Error on heldout test set}: The mean L2 error for heldout test dataset is 0.0012 for office space and 0.0011 for kitchen. We suspect that error for test images is higher due to slight overfitting on training images. We also train the student network from scratch which can result in poor learning of semantic and geometric image features. Hence, we have done some experiments with adding pretrained models as a feature extractor to the student model. As discussed in Ablation Studies, the network weights will be frozen and we will only learn the scene coordinate regression head. We compare this performance with a unfrozen pretrained model in which the weights get updated through the learning process. 

    \begin{figure}[!h]
    \centering
    \includegraphics[width=\linewidth]{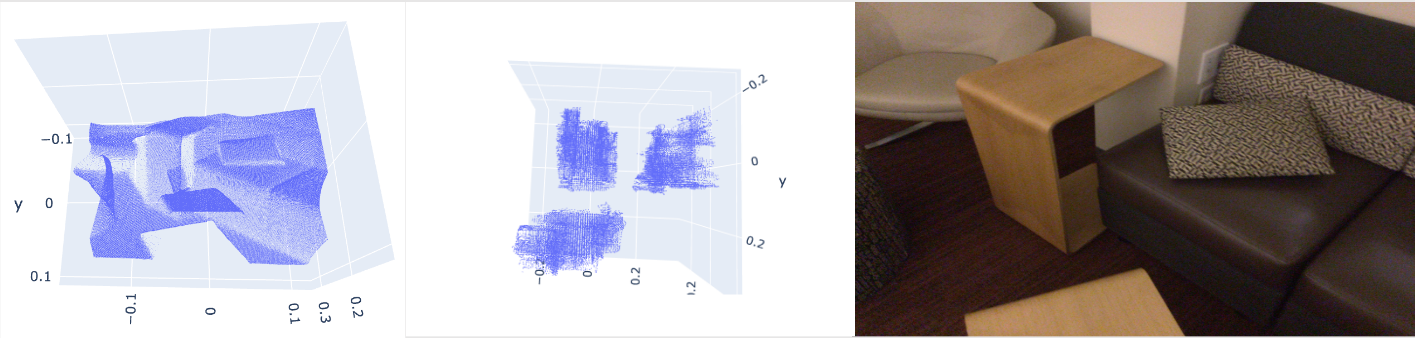}
    \caption{Visual Localization by Dust3r (left) and the Vanilla CNN model (middle), compared to the original image (right)}
    \label{fig:visual_loc_vanilla}
    \end{figure}

    \begin{figure}[!h]
    \centering
    \includegraphics[width=\linewidth]{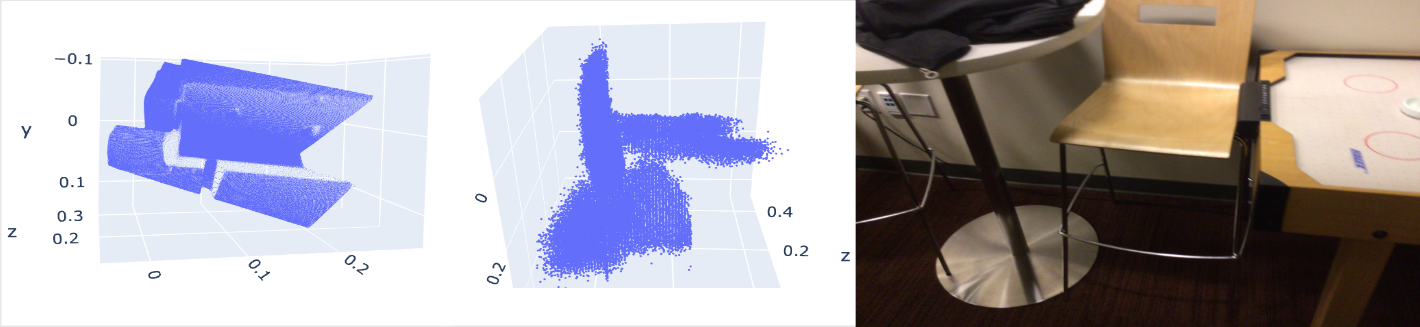}
    \caption{Visual Localization by Dust3r (left) and the MobilenetV3 model (middle), compared to the original image (right)}
    \label{fig:visual_loc_mobilenet}
    \end{figure}


\end{enumerate}
\section{Ablation Studies}

\subsection{Hyperparameter Tuning}
We perform ablation studies on tuning the hyperparameters of our student models to analyze performance improvements.
\begin{enumerate}
    \item[A.] \textbf{Epochs}: The first hyperaparameter that was varied was number of epochs of training. The model chosen was the MoboilenetV3 with a Conv Head for this study. Figure \ref{fig:hyper_epochs} shows the training losses for 3 scenes all representing the kitchen. 

    The test losses for this study are shown in Table \ref{tab:hyper_epochs}.
    \begin{figure}[!h]
    \centering
    \includegraphics[width=\linewidth]{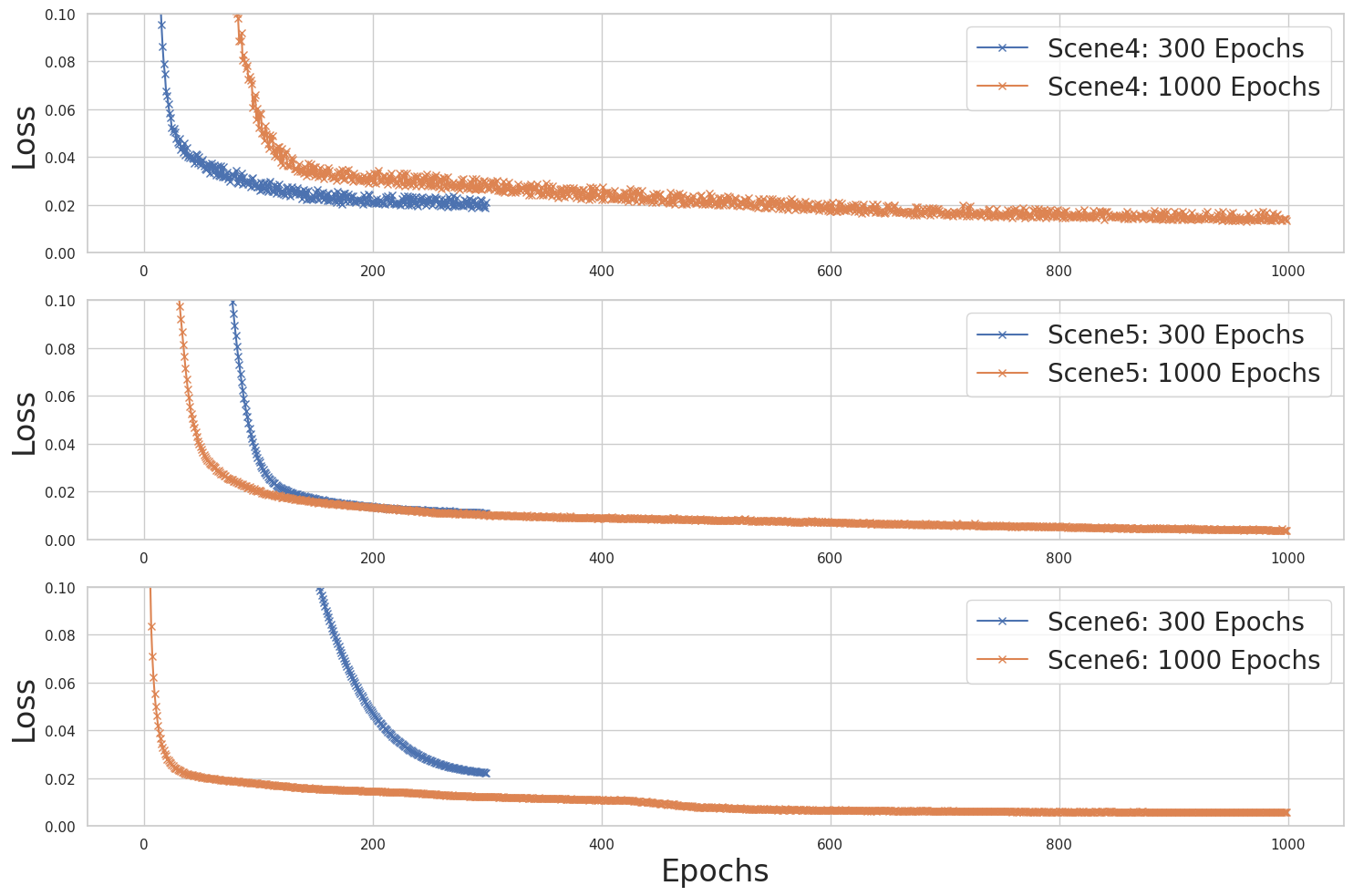}
    \caption{Comparison of Training Loss Error versus varying Epochs for 3 Comparable Scenes}
    \label{fig:hyper_epochs}
    \end{figure}

    \begin{table}[]
    \centering
    \begin{tabular}{||c | c | c ||}
         \hline
         Scenes & 300 Epochs & 1000 Epochs \\ [0.5ex] 
         \hline\hline
         Scene 4 & 3.92e-03 & \textbf{3.44e-03} \\ 
         \hline
         Scene 5 & 3.30e-03 & \textbf{2.21e-03} \\
         \hline
         Scene 6 & 4.54e-03 & \textbf{1.78e-03} \\
         \hline
    \end{tabular}
    \caption{Comparison of Average Test Errors for varying the number of training epochs of the student model between 300 and 1000 epochs.z}
    \label{tab:hyper_epochs}
    \end{table}

    We can observe from the table and figures that our training losses and test losses over different scenes are quite similar when trained under 300 epochs, indicating that the model has not overfit for each scene. However, it does not justify whether the model is robust as it could be underfitting. Thus, we test this hypothesis by training the model for a higher number of epochs (1000). We can notice a decrease in our loss values for both train and test data for most scenes, while in some scenese such as Scene 4, the losses do not differ much from the case of 300 epochs. In conclusion, training for higher number of epochs leads to better results for our model.

    \item[B.] \textbf{Pre-trained Weights}: The next hyperparameter that was tuned was the pre-trained weights of the Mobilenet model. Specifically, a comparison of the weights staying frozen during training or made unfrozen and updated during training was analyzed. Similar to the epochs study, Figure \ref{fig:hyper_frozen} shows the training losses for three more scenes that represent the kitchen environment and the test losses for this study are shown in Table \ref{tab:hyper_frozen}.
    \begin{figure}[!h]
    \centering
    \includegraphics[width=\linewidth]{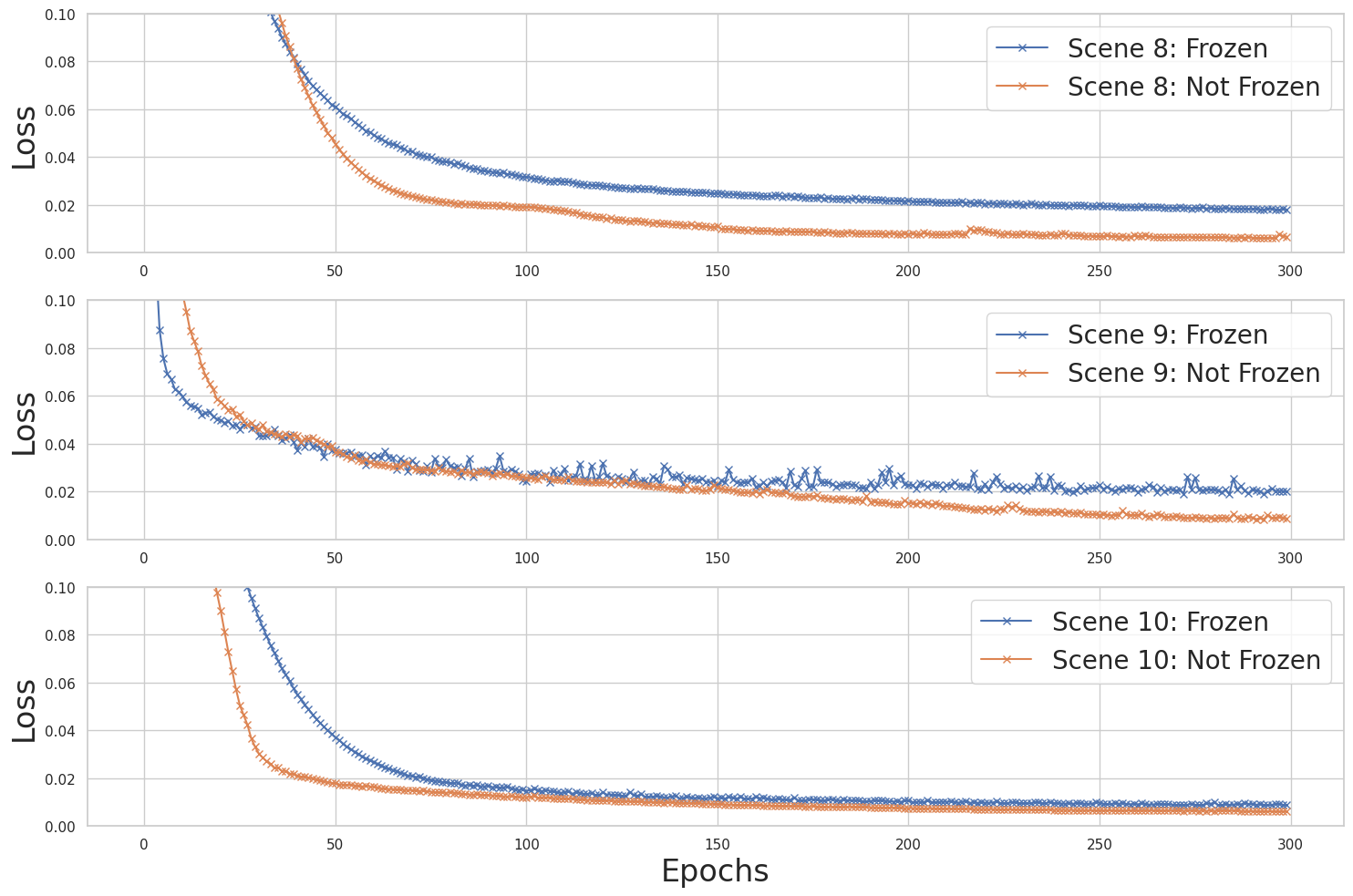}
    \caption{Comparison of Training Loss Error for a Frozen pre-trained model versus Non-Frozen for 3 Comparable Scenes}
    \label{fig:hyper_frozen}
    \end{figure}

    \begin{table}[]
    \centering
    \begin{tabular}{||c | c | c ||}
         \hline
         Scenes & Frozen Weights & Unfrozen Weights \\ [0.5ex] 
         \hline\hline
         Scene 8 & 4.23e-03 & \textbf{1.98e-03} \\ 
         \hline
         Scene 9 & \textbf{9.73e-03} & 1.10e-02 \\
         \hline
         Scene 10 & 3.39e-03 & \textbf{1.82e-03} \\
         \hline
    \end{tabular}
    \caption{Comparison of Average Test Errors when freezing the pre-trained model weights versus updating them during training}
    \label{tab:hyper_frozen}
    \end{table}

    From this study, we notice that unfreezing the weights and letting the Mobilenet model learn more has a significant impact on the training and test losses. The training losses are much lower for the unfrozen weight model and the test losses are comparable, if not significantly lower. We can conclude by saying that it is better to let the pre-trained model update its weights to learn scene-specific information rather than purely relying on its existing feature representations.

    \item[C.] \textbf{ViT Hyperparameters}:
    The Vision Transformer model was tested with:
    \begin{itemize}
        \item Patch size: 16, 32, 64
        \item Number of encoder/decoder blocks: 4, 6, 8
        \item Number of heads: 4, 8
        \item Latent dimensions: 64, 128, 256
    \end{itemize} 

    If patch sizes were too small (e.g., 16), the features were too local, resulting in more artifacts as seen in Figure \ref{fig:patch16}
    
    . By increasing the patch size, convergence became more stable and optimal (approximately 1.3 cm error).
    \begin{figure}
        \centering
        \begin{subfigure}[b]{0.2\textwidth}
            \includegraphics[scale=0.2]{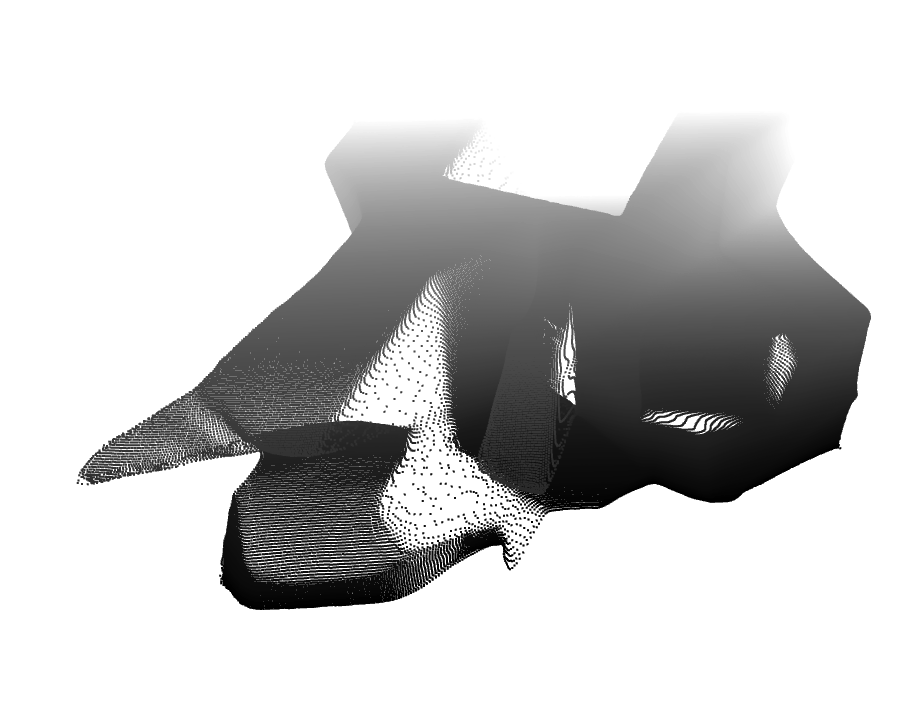}
            \caption{Ground Truth}
        \end{subfigure}
        \begin{subfigure}[b]{0.2\textwidth}
            \includegraphics[scale=0.2]{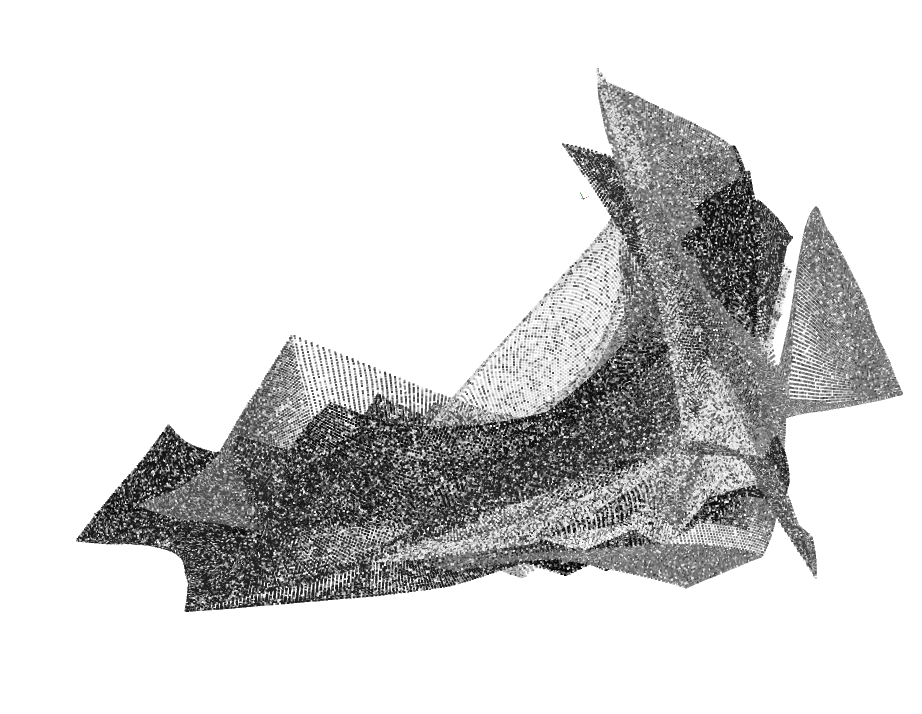}
            \caption{Output}
        \end{subfigure}
        \caption{Decreasing patch size increases artifacts}
        \label{fig:patch16}
    \end{figure}
    \\ 
    Increasing the number of encoder/decoder blocks did not necessarily improve performance. In fact, it was detrimental, as the network became too deep, and the limited number of training images was insufficient for learning the 3D structure effectively, leading to underfitting. The number of heads exhibited the same behavior. This can be seen through the losses/convergence curves in Figure \ref{fig:loss_comp_encoder_number}

    \begin{figure}[h]
        \centering
        \includegraphics[width=\linewidth]{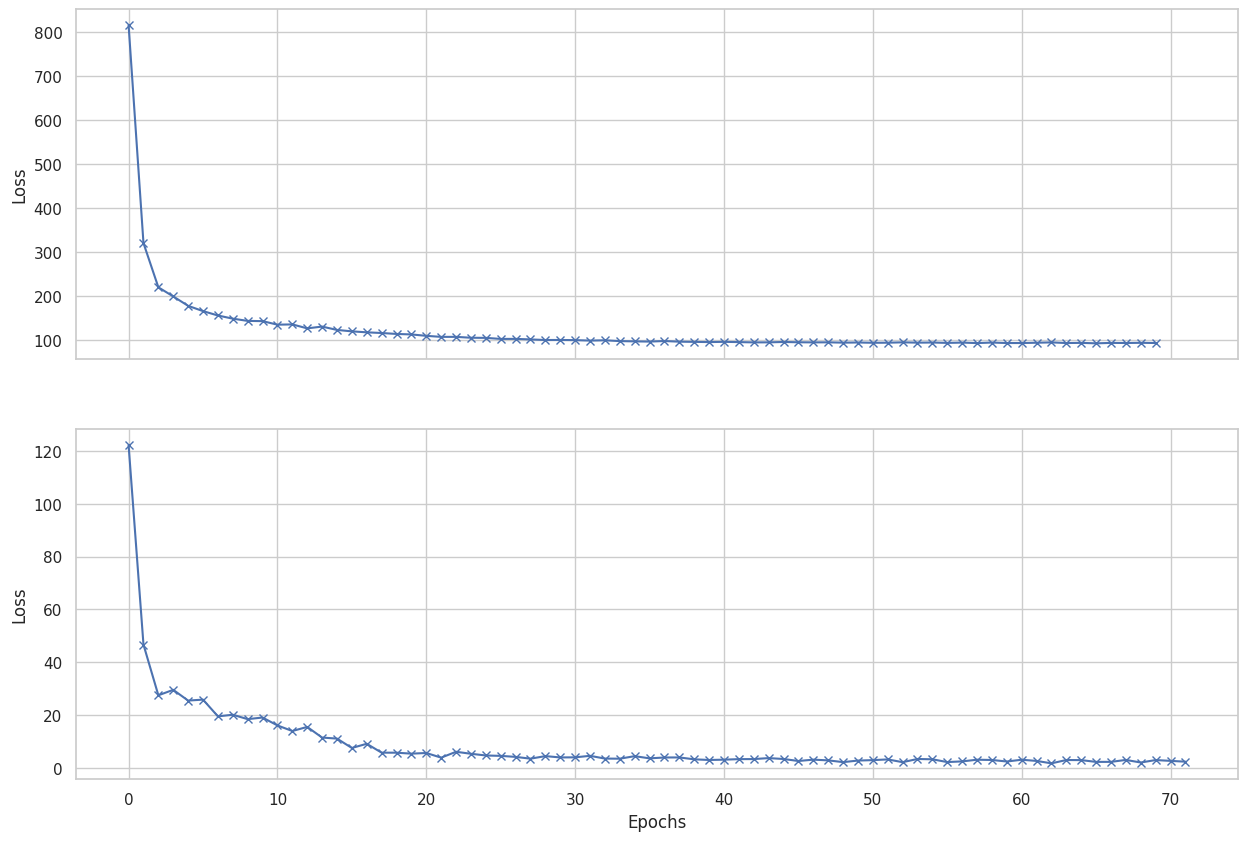}
        \caption{12 (left), 4 (right) encoder/decoder blocks. ViT with more blocks and deeper architecture converges to less optimal value and slower.}
        \label{fig:loss_comp_encoder_number}
    \end{figure}

    Increasing the latent dimensions enhanced generation capability in terms of features without making the network too deep, thereby not hindering loss convergence significantly. Results for 256 dimensions are shown in Figure \ref{fig:latent}

    \begin{figure}
        \centering
        \begin{subfigure}[b]{0.2\textwidth}
            \includegraphics[scale=0.2]{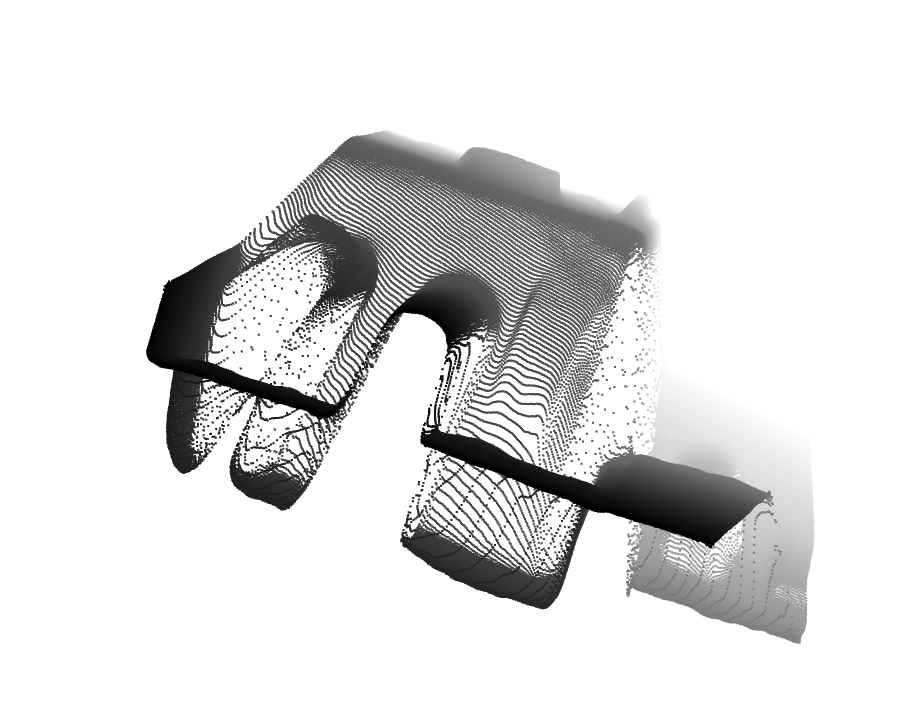}
            \caption{Ground Truth}
        \end{subfigure}
        \begin{subfigure}[b]{0.2\textwidth}
            \includegraphics[scale=0.2]{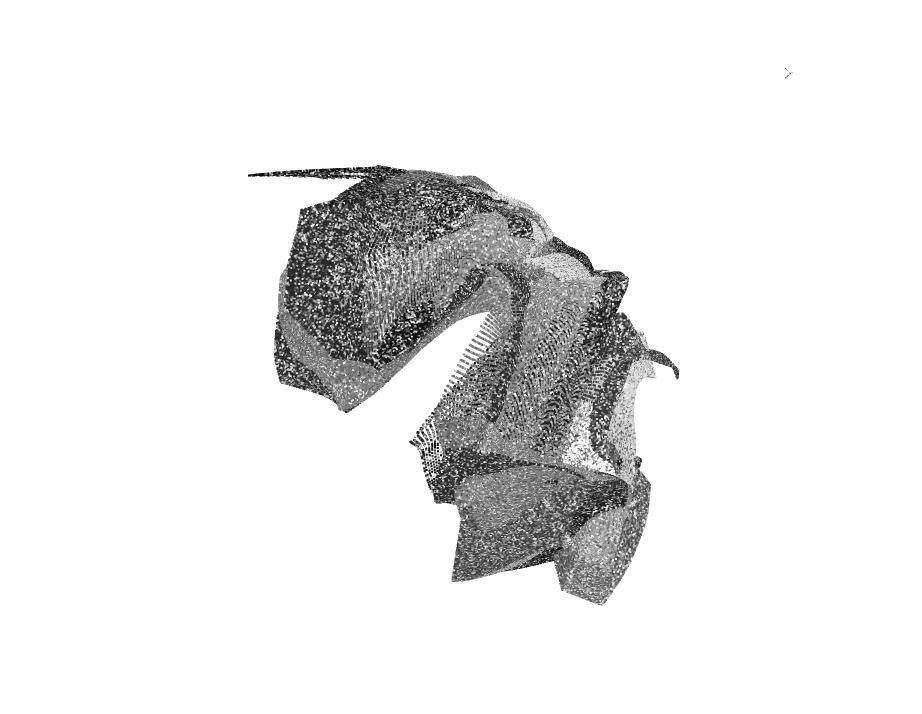}
            \caption{Output}
        \end{subfigure}
        \caption{Latent dimension 256: Folds are seen and features are learnt}
        \label{fig:latent}
    \end{figure}

    Finally, the best model was chosen with hyperparameters: 200 epochs, 256 latent dimensions, 6 encoder/decoder blocks and 4 attention heads. The results are presented in the next section for our best model.
\end{enumerate}
\section{Conclusion}

    \begin{figure}
        \centering
        \begin{subfigure}[b]{0.2\textwidth}
            \includegraphics[scale=0.2]{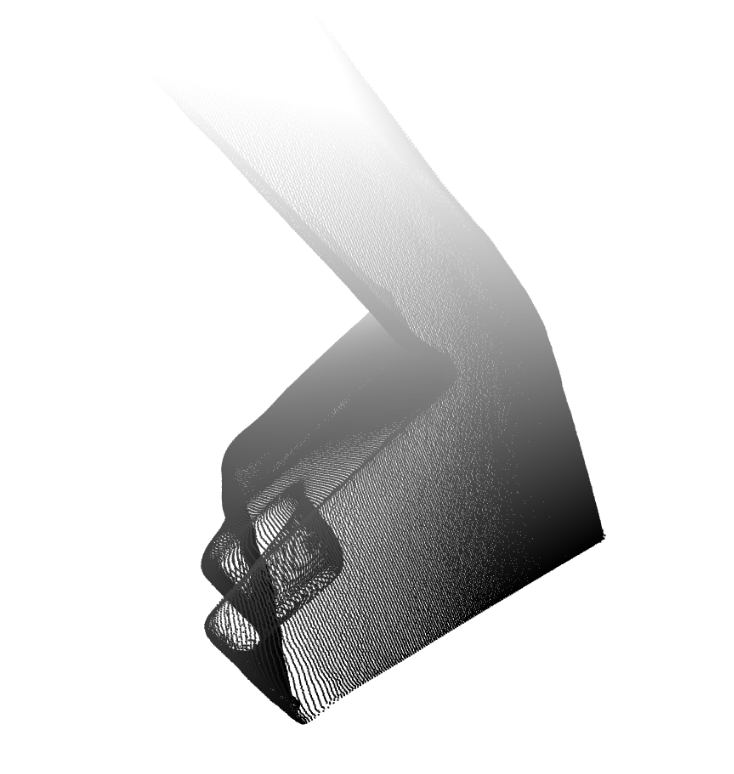}
            \caption{Ground Truth: Apt 2 scene}
        \end{subfigure}
        \begin{subfigure}[b]{0.2\textwidth}
            \includegraphics[scale=0.2]{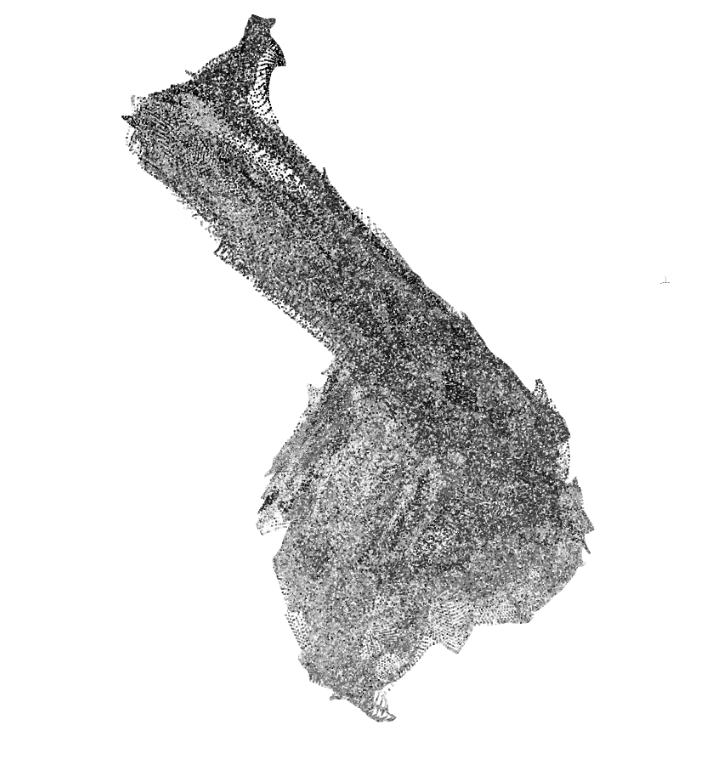}
            \caption{ViT Output: Apt 2 scene}
        \end{subfigure}
        \begin{subfigure}[b]{0.2\textwidth}
            \includegraphics[scale=0.2]{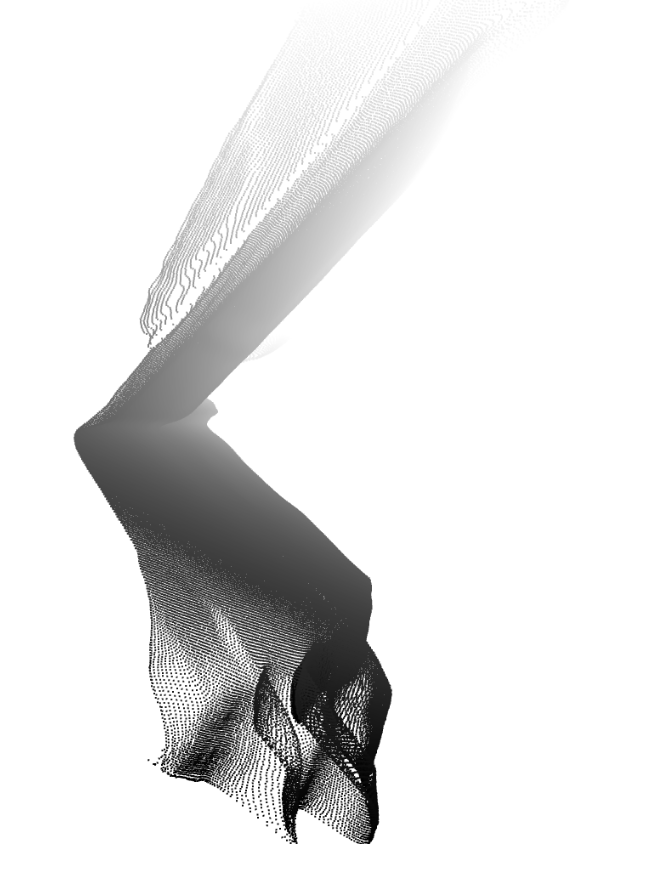}
            \caption{Ground Truth: Living Room 1 scene}
        \end{subfigure}
        \begin{subfigure}[b]{0.2\textwidth}
            \includegraphics[scale=0.2]{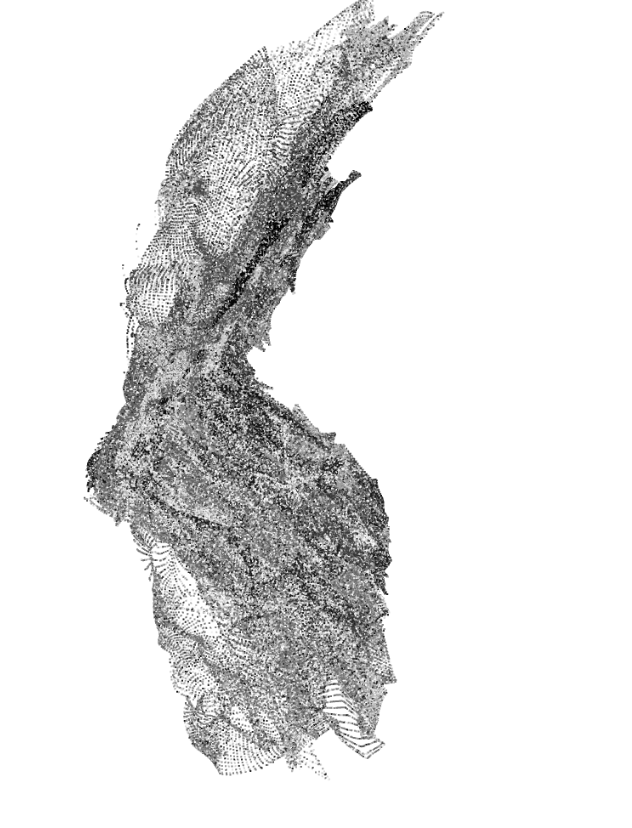}
            \caption{ViT Output: Living Room 1 scene}
        \end{subfigure}

        \caption{ViT Output of 2 couch scenes - taken from different angles}
        \label{fig:best-model-1}
    \end{figure}

In conclusion, we observe that the Vision Transformer produces 3D reconstructions that are comparable to Dust3r. The Vanilla and pre-trained Mobilenet models had several issues. They were only able to reconstruct some objects in the scene but were unable to reconstruct planes such as walls, floor, etc. By comparison, the ViT model is able to reconstruct full scene.

Overall, through our detailed investigations, we compared loss values between vanilla CNN and a pretrained mobilenet model backbone. We also compared the performance of frozen and unfrozen MobileNet weights and further decreased loss by using Vision Transformer model, which is also our best trained model. We demonstrate superior performance of the ViT model is Figure \ref{fig:latent} and \ref{fig:best-model-1}. We would like to note that all our models are in the range of 5-45MB, which is much smaller than the original Pretrained Dust3R model that is 2.2GB in size. Hence, we conclude that building a smaller light-weight network for scene-specific vision tasks is compute friendly.

For our future work, we would like to further improve the vision transformer model and make the predicted point cloud surfaces smoother. We would also like to apply our small and scene-specific trained network to perform downstream tasks such as Localization/Visual Slam or another down-stream task highlighted in \cite{wang2023dust3r}.

\bibliographystyle{ieee_fullname}

\end{document}